\documentclass[journal]{IEEEtran}

\ifCLASSINFOpdf
\usepackage[pdftex]{graphicx}
\graphicspath{{./}}

\DeclareGraphicsExtensions{.png,.pdf,.jpg,.jpeg}
\else
   \usepackage[dvips]{graphicx}
   \graphicspath{{Figures/}}
     \DeclareGraphicsExtensions{.png,.pdf,.jpg,.jpeg}
\fi

\usepackage{arydshln} 
\usepackage{fancyhdr}

\usepackage{algorithmic}
\usepackage{algorithm}
\usepackage{array}
\usepackage[caption=false,font=normalsize,labelfont=sf,textfont=sf]{subfig}
\usepackage{textcomp}
\usepackage{stfloats}
\usepackage{url}
\usepackage{verbatim}
\usepackage{cite}
\usepackage{multirow}
\usepackage{pifont}
\usepackage{amsmath,amssymb,amsfonts,amsbsy}
\usepackage{bm}
\usepackage{graphicx}
\usepackage{mathtools}
\usepackage{array,ragged2e}
\usepackage{booktabs}
\usepackage{makecell}

\usepackage{color}
\usepackage[table,xcdraw,dvipsnames]{xcolor}
\definecolor{myPink}{rgb}{0.9294, 0.0078, 0.5490}
\definecolor{Gray}{gray}{0.92}
\definecolor{my_color}{HTML}{E8F3F1}
\definecolor{my_color1}{HTML}{FFEACE}
\definecolor{my_color2}{HTML}{FBEAFF}
\definecolor{my_color3}{HTML}{FFC1B5}
\usepackage[
  colorlinks=true,
  linkcolor=blue,  
  citecolor=blue,  
  urlcolor=myPink,   
  linkbordercolor={1 1 1} 
]{hyperref}
\usepackage{threeparttable}
\usepackage{diagbox}
\usepackage{orcidlink}
\usepackage{pifont}

\newcommand{\cmark}{\ding{51}}
\newcommand{\xmark}{\ding{55}}

\makeatletter
\let\NAT@parse\undefined
\makeatother

\begin{document}
 
\title{DVANet: Degradation-aware Visual-prior Alignment Network for Image Restoration} 

\author{
        Yanjie Tu,
        Qingsen Yan, 
        Axi Niu,
        Tao Hu, 
        Haokui Zhang, and 
        Jiantao Zhou

\thanks{This work is supported by the NSFC of China under Grant 62301432 and 62306240.
\emph{(Corresponding author: Qingsen Yan.)}}%

\thanks{Yanjie Tu, Qingsen Yan, Axi Niu, Tao Hu, and Haokui Zhang are with the School of Computer Science, Northwestern Polytechnical University, Xi'an 710000, China, and Qingsen Yan is also with the Shenzhen Research Institute of Northwestern Polytechnical University, Shenzhen 518057, China (e-mail: yanjietu@mail.nwpu.edu.cn; qingsenyan@nwpu.edu.cn; nax@nwpu.edu.cn; taohu@mail.nwpu.edu.cn; hkzhang@nwpu.edu.cn).}
\thanks{Jiantao Zhou is with the State Key Laboratory of Internet of Things for Smart City, University of Macau, Macau 999078, China (e-mail: jtzhou@um.edu.mo).}

}

\markboth{Journal of \LaTeX\ Class Files,~Vol.~14, No.~8, August~2021}%
{Shell \MakeLowercase{\textit{\textit{et al.}}}: A Sample Article Using IEEEtran.cls for IEEE Journals}

\maketitle

%
\begin{abstract}
All-in-One image restoration aims to develop a unified restoration framework for handling diverse degradation types. Existing end-to-end methods usually regard the restoration process as a black-box mapping, lacking an explicit optimization interpretation. Although deep unfolding provides an interpretable iterative modeling paradigm for image restoration, existing methods mostly rely on fixed degradation assumptions or predefined degradation information, making them difficult to adapt to unified restoration requirements under complex degradations and locally damaged content. This limitation restricts their performance in degradation suppression and structural detail recovery. To address these issues, this paper proposes DVANet, a deep unfolding network inspired by the half-quadratic splitting optimization algorithm, which formulates unified image restoration under complex degradations as a collaborative unfolding process between degradation-aware observation consistency and visual-prior-guided reconstruction. Specifically, in the degradation-aware observation consistency branch, a degradation representation module is employed to extract global degradation attributes and local degradation cues, and degradation-conditioned mapping is used to enhance the model's adaptability to different degradation types. In the visual-prior-guided reconstruction branch, DINOv3 is introduced to provide structural and semantic information as hierarchical visual priors, thereby complementing the missing structural information in damaged regions and improving detail recovery. Extensive experiments demonstrate that DVANet achieves superior or competitive performance on multi-scenario degradation and cross-domain image restoration tasks, showing favorable degradation adaptability and generalization ability. The source code is available at \href{https://github.com/leoyjTu/DVANet}{DVANet}.

\end{abstract}

\IEEEpeerreviewmaketitle
\section{Introduction} 
\IEEEPARstart{I}{mage} restoration (IR) is a fundamental problem in low-level vision, aiming to recover a high-quality latent clean image from a degraded observation. In real-world imaging processes, image quality is often affected by sensor noise, motion blur, and adverse weather conditions such as rain, snow, and haze, as well as low-light imaging environments. These degradations not only reduce the visual quality and usability of images, but also further compromise the reliability of high-level vision tasks, such as object detection and scene understanding~\cite{baryir,bioir}. Recently, convolutional neural network- and transformer-based methods have achieved remarkable progress in image denoising~\cite{denoising}, deblurring~\cite{deblurring}, deraining~\cite{drsformer}, dehazing~\cite{dehazeformer}, and low-light enhancement~\cite{low-light}. However, most existing methods are still designed and trained for specific degradation types. When facing complex, diverse, or even composite degradations in real-world scenarios, their restoration performance and generalization ability are often difficult to guarantee~\cite{ir-sde,fourmer,convir,mb-taylorformerv2}. Therefore, All-in-One image restoration has gradually become an important research direction in this field, aiming to achieve robust restoration of multiple degradation types within a unified framework.

\begin{figure}[tbp]   %
	\centerline{\includegraphics[page=1,trim = 0mm 0mm 0mm 0mm, clip, width=1\linewidth]{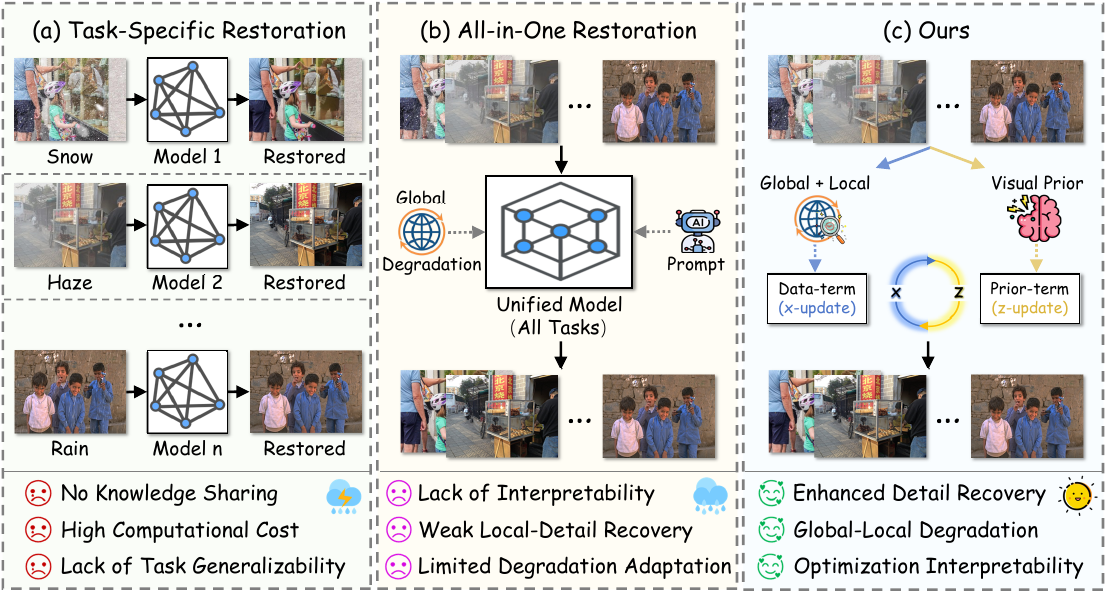}}
\vspace{-0.3em}
	\caption{Conceptual comparison of task-specific restoration, existing All-in-One restoration methods, and the proposed degradation-aware visual-prior unfolding framework. The proposed framework improves degradation adaptability and structural detail recovery by jointly modeling observation consistency and visual-prior-guided reconstruction in an interpretable unfolding manner.}
	\label{fig:motivation}
	\vspace{-1em}
\end{figure}

From a modeling perspective, image restoration can generally be formulated as an ill-posed inverse problem, which aims to estimate the latent clean image from a degraded observation. In general, the degradation model can be written as
\begin{equation}
y = \mathcal{A}(x) + n,
\label{eq:degradation_model}
\end{equation}
where $x$ denotes the unknown latent clean image, $y$ represents the degraded observation, $\mathcal{A}(\cdot)$ is the degradation operator, and $n$ denotes the noise term. Since the degradation process is usually accompanied by information loss, noise perturbation, and structural corruption, relying solely on the observed image makes it difficult to recover stable and plausible results. Therefore, classical image restoration methods usually introduce prior constraints and formulate the restoration process as the following regularized optimization problem:
\begin{equation}
\hat{x} = \arg\min_{x} \frac{1}{2}\|\mathcal{A}(x)-y\|_2^2 + \lambda \Psi(x),
\label{eq:regularized_optimization}
\end{equation}
where the first term is the data fidelity term, which enforces consistency between the restored result and the observed image; the second term denotes the image prior or regularization term; and $\lambda$ is a weighting parameter that balances data consistency and prior constraints. This formulation indicates that stable image restoration usually requires a proper balance between observation consistency constraints and image prior modeling.

Based on the above inverse problem, existing image restoration methods can be roughly divided into model-based methods, deep learning-based methods, and deep unfolding methods that combine the advantages of both~\cite{usrnet,dgunet,meinhardt2017learning,vlu-net}. Model-based methods usually exhibit good physical interpretability, but they rely on manually designed priors and thus have limited representation capability in complex real-world degradation scenarios~\cite{dabov2007image,dong2011image}. Deep learning-based methods can learn powerful nonlinear mappings from large-scale data, but their end-to-end black-box mapping paradigm often lacks explicit optimization interpretation and degradation constraints~\cite{airnet,promptir}. In recent years, deep unfolding networks (DUNs) have bridged model interpretability and deep representation learning by unfolding classical optimization algorithms into trainable networks~\cite{dgunet,usrnet,dong2018denoising}. However, most existing deep unfolding methods are designed for specific restoration tasks and usually depend on fixed degradation assumptions or predefined degradation information, making them difficult to adapt to complex and diverse degradation scenarios with locally damaged content in real-world applications.

Half-quadratic splitting (HQS)~\cite{hqs} provides a naturally structured modeling perspective for the above problem. This formulation decomposes the restoration process into two mutually collaborative subproblems: one focuses on observation consistency modeling for the data variable $x$, while the other emphasizes image prior recovery for the prior variable $z$. For a known degradation operator, the update of the data variable usually relies on $\mathcal{A}$ and its adjoint operator $\mathcal{A}^{\mathrm{T}}$. However, in All-in-One image restoration, degradation types are diverse and complex, making it difficult to explicitly construct a reliable degradation operator. Therefore, how to learn degradation-adaptive observation consistency modeling within a unified unfolding framework, while simultaneously leveraging hierarchical visual priors to enhance the recovery of structural details in locally damaged or severely degraded regions, is the core problem addressed in this paper.

Based on the above analysis, this paper proposes DVANet, a deep unfolding network for unified restoration under complex degradations. Taking HQS-based~\cite{hqs} variable splitting as the theoretical prototype, DVANet formulates image restoration in the multi-scale feature space as a collaborative unfolding process between degradation-aware observation consistency and visual-prior-guided reconstruction. Specifically, in the degradation-aware observation consistency branch, DVANet employs a global-local joint degradation representation module to extract image-level global degradation attributes and region-level local degradation cues from the input image and further utilizes degradation-conditioned data mapping to enhance the model's adaptive modeling capability for different degradations. In the visual-prior-guided reconstruction branch, DVANet introduces a frozen DINOv3~\cite{dinov3} to provide hierarchical visual priors. Through a lightweight prior adapter and a residual gated modulation mechanism, hierarchical visual features are injected into the update process of the prior variable, thereby enhancing the recovery of structural details in locally damaged or severely degraded regions.

Our contributions can be summarized as follows:

\begin{itemize}
    \item We propose DVANet, a deep unfolding network for unified restoration under complex degradations. DVANet formulates image restoration as a collaborative unfolding process between degradation-aware observation consistency and visual-prior-guided reconstruction, thereby improving interpretability and generalization ability in complex degradation scenarios.

    \item We construct a degradation-aware observation consistency mechanism and a visual-prior-guided reconstruction mechanism. The former enhances complex degradation modeling through global-local degradation representation and degradation-conditioned data mapping, while the latter leverages hierarchical visual priors from DINOv3 to improve structural detail recovery.

    \item Extensive experiments demonstrate that DVANet achieves superior or competitive performance on single degradation, complex nighttime, composite degradation, and cross-domain image restoration tasks, with consistent improvements in both quantitative and qualitative evaluations.
\end{itemize}
%

\section{Related Work} 

\subsection{All-in-One Restoration} 
All-in-One Image Restoration (AiOIR) aims to handle multiple image degradations using a unified model and has become an important research direction in low-level vision in recent years~\cite{airnet,idr,vivnet,baryir}. Compared with training separate models for different degradation tasks, AiOIR has clear advantages in model storage, deployment efficiency, and task scalability. Therefore, it is regarded as an important direction for addressing diverse degradation types, complex degradation forms, and the coexistence of multiple degradations in real-world scenarios.

Existing AiOIR methods mainly focus on degradation awareness, task adaptation, and the utilization of visual priors. Early methods usually alleviate conflicts caused by multi-task parameter sharing through degradation-representation learning. For example, AirNet~\cite{airnet} employs contrastive learning to obtain discriminative degradation embeddings. Subsequently, methods such as PromptIR~\cite{promptir}, ProRes~\cite{prores}, and NDR-Restore~\cite{ndr} introduce visual prompts, scene descriptors, or degradation queries to conditionally modulate the restoration process. In addition, methods such as InstructIR~\cite{instructir} and UniProcessor~\cite{uniProcessor} further leverage natural language or multimodal prompts to enhance the model's understanding of complex restoration requirements. Recently, MoCE-IR~\cite{moceir} improved task adaptation capability through an expert activation mechanism, while DA-CLIP~\cite{da-clip}, DINO-IR~\cite{dino-ir}, and related methods exploit visual representations or semantic contexts provided by pre-trained models to enhance structure-aware restoration capability.

Although existing AiOIR methods have achieved significant progress in unified modeling of multiple degradations, most of them still mainly rely on end-to-end deep mappings to learn the restoration relationship from degraded images to clean images. As a result, it is difficult to explicitly associate the restoration process with observation consistency modeling and image prior constraints, leading to a lack of clear optimization interpretation. In complex degradation scenarios or scenes with locally damaged content, such a black-box restoration process further limits the interpretability and controllability of model behavior. To this end, Deep Unfolding Networks provide a more interpretable modeling paradigm for unified image restoration by unfolding traditional optimization processes into trainable networks.

\subsection{Deep Unfolding Networks} 
Deep Unfolding Networks (DUNs) unfold traditional iterative optimization algorithms into trainable network stages, thereby introducing the representation capability of deep networks while preserving optimization interpretability~\cite{dgunet,deepsn-net}. This idea has been widely applied to inverse problem tasks such as compressive sensing, image super-resolution, and image restoration. Typical methods are usually based on optimization algorithms such as ISTA~\cite{ista}, PGD~\cite{pgd}, ADMM~\cite{admm}, or HQS~\cite{hqs} and map iterative update steps into end-to-end trainable network modules~\cite{deepsn-net,vlu-net}.

In the field of image restoration, existing DUN methods usually decompose the restoration problem into an observation consistency subproblem and a prior regularization subproblem through variable splitting or proximal optimization. Specifically, the observation consistency subproblem constrains the consistency between the restored result and the degraded observation, while the prior subproblem approximates the traditional regularization term or proximal operator using learnable modules such as CNNs, U-Nets, and Transformers. For example, methods such as DCDicL~\cite{dcdicl}, USRNet~\cite{usrnet}, and DGUNet~\cite{dgunet} combine explicit optimization updates with deep prior modules, achieving favorable performance in specific restoration tasks. 

However, most existing DUN methods are designed for specific degradation tasks and rely on manually defined degradation models or explicit degradation parameters, making them difficult to directly adapt to unified restoration scenarios with diverse and complex degradation types~\cite{dong2018denoising,usrnet}. In addition, their prior modules are mostly modeled as task-related learnable denoisers or restorers, with limited exploration of hierarchical priors provided by vision foundation models for enhancing structural detail recovery. In contrast, DVANet models degradation-aware observation consistency and visual-prior-guided reconstruction within an HQS-inspired dual-variable unfolding framework, thereby better meeting the requirements of unified image restoration in complex degradation scenarios.

\begin{figure*}[!htp]  
	\centerline{\includegraphics[page=1,trim = 0mm 0mm 0mm 0mm, clip, width=1\linewidth]{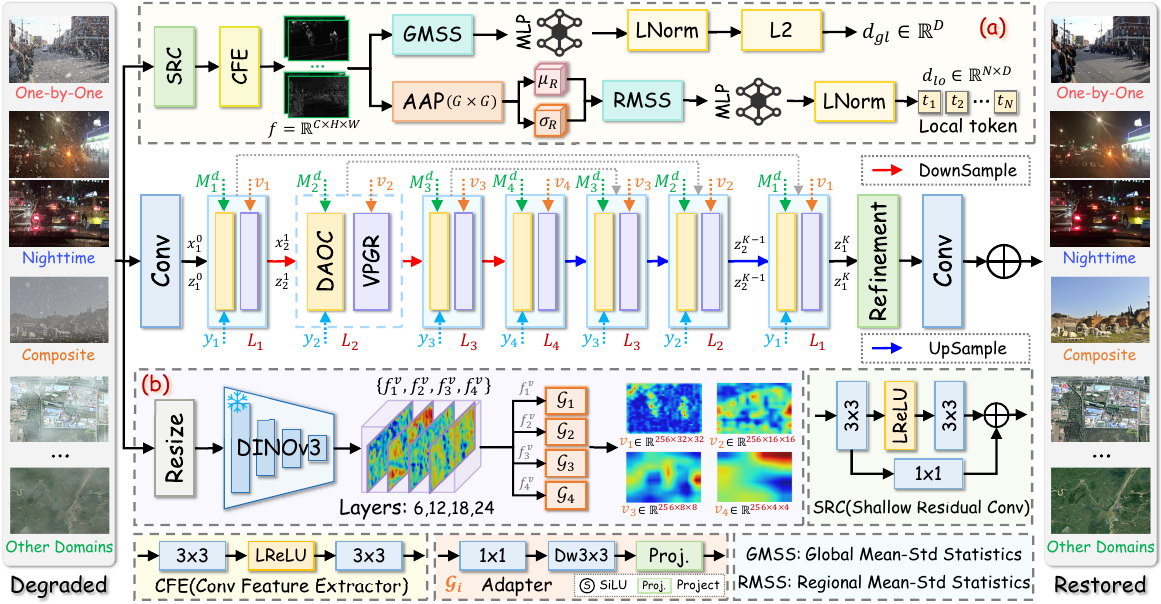}}
\vspace{-0.3em} 
	\caption{Overall architecture of DVANet. Given a degraded image, DVANet extracts two types of auxiliary cues: (a) global-local degradation representations from the degradation representation block, and (b) hierarchical visual priors from the frozen DINOv3 encoder with lightweight prior adapters. These cues are then used to guide the degradation-aware observation consistency update and visual-prior-guided reconstruction in the dual-variable unfolding process.}  
	\label{fig:overall}
	\vspace{-1em}
\end{figure*}
%

\section{Method} 
\label{Method}
\subsection{Overview Pipeline} 
As shown in Figure~\ref{fig:overall}, given an input degraded image $I_L$, DVANet first extracts two types of auxiliary information from the input image: degradation-related information and hierarchical visual priors. Specifically, the degradation representation module extracts an image-level global degradation vector $d_{gl}$ and region-level local degradation tokens $d_{lo}=\{d_{lo,i}\}_{i=1}^{N}$ from $I_L$, which are used to describe the overall degradation attributes and local degradation cues, respectively. Meanwhile, the frozen DINOv3~\cite{dinov3} visual encoder extracts multi-level visual features, which are further transformed into multi-scale visual prior features $V=\{v_l\}_{l=1}^{L}$ through a lightweight prior adapter. Subsequently, the input degraded image is mapped into the initial data variable $x_1^{(0)}$, the initial prior variable $z_1^{(0)}$, and the observation feature $y_1$, respectively. Among them, the data variable $x$ is used to model degradation observation constraints and observation consistency, the prior variable $z$ is used to model image priors and integrate hierarchical visual priors, and the observation feature $y$ serves as the observation reference in the data variable update. DVANet embeds the dual-variable update into a multi-scale encoder-decoder pathway, enabling the restoration process to progressively accomplish degradation suppression and structural detail recovery at different feature scales.

Let $l$ denote the current feature scale and $k$ denote the unfolding stage index along the multi-scale encoder-decoder pathway. The data variable, prior variable, and observation feature at the $l$-th scale in the $k$-th unfolding stage are denoted as $x_l^{(k)}$, $z_l^{(k)}$, and $y_l$, respectively, and the visual prior feature at the corresponding scale is denoted as $v_l$. In each unfolding stage, DVANet sequentially performs degradation-aware observation consistency update and visual-prior-guided reconstruction.

First, the data variable is updated through the degradation-aware observation consistency branch:
\begin{equation}
\small
\begin{aligned}
x_l^{(k+1)} = x_l^{(k)} - \tau_l^{(k)}
\Big(
&T_l\left(\Phi_l^d\left(x_l^{(k)}; d_{gl}, d_{lo}\right)-y_l\right) \\
&+ \mu_l^{(k)}\big(x_l^{(k)}-z_l^{(k)}\big)
\Big),
\end{aligned}
\label{eq:data_update}
\end{equation}
where $\Phi_l^d(\cdot; d_{gl}, d_{lo})$ denotes the degradation-conditioned data mapping, which is used to model degradation-related data responses; $T_l(\cdot)$ represents a learnable adjoint-like residual mapping; and $\tau_l^{(k)}$ and $\mu_l^{(k)}$ denote the learnable step size and the dual-variable coupling weight, respectively.

Subsequently, the visual-prior-guided reconstruction branch predicts the prior variable according to the updated data variable and the DINOv3 visual prior at the corresponding scale:
\begin{equation}
    \tilde{z}_l^{(k+1)} = P_l\left(x_l^{(k+1)}; v_l\right),
    \label{eq:prior_prediction}
\end{equation}
where $P_l(\cdot; v_l)$ denotes the reconstruction operator modulated by the visual prior. To avoid the instability caused by directly replacing the prior variable, DVANet adopts a residual strategy to update the prior variable:
\begin{equation}
    z_l^{(k+1)} = z_l^{(k)} + \alpha_l^{(k)}
    \left(\tilde{z}_l^{(k+1)} - z_l^{(k)}\right),
    \label{eq:prior_update}
\end{equation}
where $\alpha_l^{(k)}$ denotes a learnable update weight that controls the update magnitude of the predicted prior variable with respect to the current prior variable. After consecutive dual-variable updates along the multi-scale encoder-decoder pathway, DVANet uses the prior variable $z_1^{(K)}$ from the final stage to generate the restoration residual and obtains the final restored image through a global residual connection:
\begin{equation}
    \hat{I} = \mathcal{R}_{out}(z_1^{(K)}) + I_L
    \label{eq:global_residual}
\end{equation}
where $\mathcal{R}_{out}$ denotes the output reconstruction layer and $K$ represents the index of the last unfolding stage in the multi-scale encoder-decoder pathway.
%

\subsection{Degradation-Aware Observation Consistency} 
In conventional unfolding methods, the update of the data variable usually relies on an explicit degradation operator and its adjoint operator. However, in unified restoration scenarios, degradation types are diverse and difficult to predefine, making it challenging to construct a reliable physical degradation operator. To address this issue, we design a degradation-aware observation consistency branch, which conditionally models the observation consistency update of the data variable using input-dependent degradation representations, enabling the model to adaptively adjust the observation consistency process.

\subsubsection{Global-Local Degradation Representation} 
To capture both image-level degradation attributes and region-level degradation cues, we design a global-local joint degradation representation module, termed the Degradation Representation Block (DRB). As shown in Figure~\ref{fig:overall}(a), given an input degraded image $I_L \in \mathbb{R}^{3 \times H \times W}$, DRB first obtains intermediate features through a shallow residual convolution module and a convolutional feature extractor:
\begin{equation}
    f = C\left(R\left(I_L\right)\right),
    \label{eq:drb_feature}
\end{equation}
where $R(\cdot)$ and $C(\cdot)$ denote the shallow residual convolution module and the convolutional feature extraction module, respectively.

Based on the feature $f$, DRB constructs degradation representations from both global and regional statistics. First, the channel-wise global mean and global standard deviation are computed along the spatial dimensions and then concatenated to obtain a global statistical vector:
\begin{equation}
    \eta_{gl} = \mathrm{Concat}\left(\mathrm{GAP}(f), \mathrm{GSP}(f)\right),
    \label{eq:global_statistics}
\end{equation}
where $\mathrm{GAP}(\cdot)$ and $\mathrm{GSP}(\cdot)$ denote global average pooling and global standard deviation pooling, respectively. Subsequently, the global degradation projection module maps $\eta_{gl}$ into the degradation embedding space, followed by normalization to obtain the image-level global degradation vector:
\begin{equation}
    d_{gl} = \mathrm{L2Norm}\left(\mathrm{LN}\left(M_{gl}\left(\eta_{gl}\right)\right)\right),
    \label{eq:global_degradation_vector}
\end{equation}
where $M_{gl}(\cdot)$ denotes the global degradation projection module, while $\mathrm{LN}(\cdot)$ and $\mathrm{L2Norm}(\cdot)$ denote layer normalization and $L_2$ normalization, respectively.

However, relying solely on the global degradation vector is insufficient to describe degradation variations across different regions within an image. Therefore, DRB further extracts local degradation tokens from region-level statistics. Specifically, the feature $f$ is adaptively aggregated into $G \times G$ regions. The total number of regions is denoted as $N=G^2$, and the local feature of the $i$-th region is denoted as $f_i$. For each region, the mean and standard deviation are computed separately to obtain the region-level statistical representation:
\begin{equation}
    \eta_{lo,i} = \mathrm{Concat}\left(\mathrm{Mean}(f_i), \mathrm{Std}(f_i)\right),
    \quad i=1,\ldots,N.
    \label{eq:local_statistics}
\end{equation}
Then, a shared local degradation projection module is used to map each regional statistic into the degradation embedding space, followed by layer normalization to obtain the local degradation tokens:
\begin{equation}
    d_{lo,i} = \mathrm{LN}\left(M_{lo}\left(\eta_{lo,i}\right)\right),
    \quad i=1,\ldots,N.
    \label{eq:local_degradation_token}
\end{equation}
Accordingly, the set of local degradation tokens is obtained as
\begin{equation}
    d_{lo} = \{d_{lo,i}\}_{i=1}^{N} \in \mathbb{R}^{N \times D},
    \label{eq:local_degradation_set}
\end{equation}
where $d_{gl}$ provides image-level degradation context, while $d_{lo}$ describes region-level degradation cues. Together, they provide multi-granularity degradation conditions for the subsequent degradation-conditioned data mapping.

\subsubsection{Degradation-Conditioned Data Mapping} 
\begin{figure}[!tbp]  
	\centerline{\includegraphics[page=1,trim = 0mm 0mm 0mm 0mm, clip, width=1\linewidth]{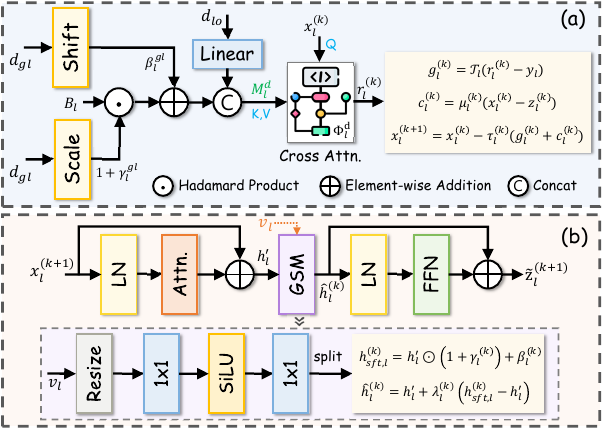}}
    \caption{(a) Degradation-conditioned data mapping guides the observation-consistency update using global and local degradation representations. (b) Visual prior-guided reconstruction injects DINOv3 priors into the prior-variable update via gated semantic modulation.}
	\label{fig:de-visual}
	\vspace{-1em}
\end{figure}

After obtaining the global degradation vector $d_{gl}$ and the local degradation tokens $d_{lo}$, we further construct a degradation-conditioned data mapping, enabling the current data variable to adaptively model degradation-related responses according to the input degradation state. As shown in Figure~\ref{fig:de-visual}(a), at the $l$-th scale, a set of learnable degradation base tokens is first introduced:
\begin{equation}
    B_l = \{b_{l,1}, b_{l,2}, \ldots, b_{l,T}\}, 
    \quad b_{l,t} \in \mathbb{R}^{C_l},
    \label{eq:base_tokens}
\end{equation}
where $T$ denotes the number of base tokens, and $C_l$ denotes the number of channels of the data variable at the $l$-th scale.

To adapt the learnable degradation tokens to the overall degradation state of the current input image, the global degradation vector is used to generate channel-wise scaling and shifting terms, which are then applied to conditionally modulate each degradation base token:
\begin{equation}
    \gamma_l^{gl} = W_{\gamma,l}(d_{gl}), 
    \quad
    \beta_l^{gl} = W_{\beta,l}(d_{gl}),
    \label{eq:global_modulation_params}
\end{equation}
\begin{equation}
    \tilde{b}_{l,t} = b_{l,t} \odot \left(1+\gamma_l^{gl}\right) + \beta_l^{gl},
    \quad t=1,\ldots,T,
    \label{eq:global_modulated_tokens}
\end{equation}
where $\odot$ denotes element-wise multiplication. Meanwhile, to incorporate region-level degradation cues, the local degradation tokens are projected into the feature space of the current scale:
\begin{equation}
    \tilde{d}_{lo,i}^{\,l} = W_{lo,l}\left(d_{lo,i}\right),
    \quad i=1,\ldots,N.
    \label{eq:local_token_projection}
\end{equation}
Subsequently, the globally modulated degradation base tokens and the local degradation tokens are concatenated and normalized to obtain the degradation-conditioned representation:
\begin{equation}
    \mathcal{M}_l^d = \mathrm{LN}\left(
    \mathrm{Concat}\left(
    \{\tilde{b}_{l,t}\}_{t=1}^{T},
    \{\tilde{d}_{lo,i}^{\,l}\}_{i=1}^{N}
    \right)
    \right),
    \label{eq:degradation_conditioned_representation}
\end{equation}
where $\mathcal{M}_l^d \in \mathbb{R}^{(T+N)\times C_l}$ denotes the degradation-conditioned representation at the $l$-th scale, which simultaneously contains input-dependent global degradation context and region-level local degradation cues.

At the $k$-th unfolding stage, we take the current data variable $x_l^{(k)}$ as the query and the degradation-conditioned representation $M_l^d$ as the key and value to construct a degradation-conditioned cross-attention mapping:
\begin{equation}
    r_l^{(k)} = \Phi_l^d\left(x_l^{(k)}; d_{gl}, d_{lo}\right)
    = \mathrm{CrossAttn}\left(x_l^{(k)}, M_l^d\right),
    \label{eq:degradation_conditioned_cross_attention}
\end{equation}
where $r_l^{(k)}$ denotes the degradation-conditioned data response and $\Phi_l^d(\cdot; d_{gl}, d_{lo})$ represents a learnable degradation-conditioned data mapping rather than a fixed physical degradation operator. Subsequently, the residual between $r_l^{(k)}$ and the observation feature $y_l$ is fed into the learnable adjoint-like residual mapping $T_l(\cdot)$ to generate the observation consistency update direction. Together with the dual-variable coupling term, the data variable is updated as
\begin{equation}
\small
    x_l^{(k+1)} = x_l^{(k)} - \tau_l^{(k)}
    \left(
    T_l\left(r_l^{(k)}-y_l\right)
    + \mu_l^{(k)}\left(x_l^{(k)}-z_l^{(k)}\right)
    \right).
    \label{eq:degradation_aware_data_update}
\end{equation}
Unlike conventional unfolding methods that rely on fixed degradation operators, this branch provides input-dependent degradation context for the observation consistency update of the data variable through global-local degradation representation and degradation-conditioned data mapping. This enables the model to adaptively adjust the observation consistency modeling process according to different degradation types, thereby enhancing its adaptive modeling capability for complex degradations.
%

%

\subsection{Visual Prior-Guided Reconstruction} 
Under severe degradation conditions, the structural and detailed information in the input image is often corrupted, making it difficult to recover reliable content by relying solely on low-level observation information. To address this issue, we introduce a frozen DINOv3~\cite{dinov3} visual encoder, which leverages its multi-level structural and semantic representations learned from large-scale pre-training as hierarchical visual priors and aligns them with multi-scale restoration features through a lightweight prior adapter.

\subsubsection{Visual Prior Adaptation} 
As shown in Figure~\ref{fig:overall}(b), given an input degraded image $I_L$, we first resample it to the input resolution required by DINOv3 and perform normalization preprocessing. The processed input is denoted as $I_{\mathrm{din}}$. Then, $I_{\mathrm{din}}$ is fed into the frozen DINOv3 encoder, and intermediate visual features are extracted from multiple specified Transformer layers:
\begin{equation}
    \{f_1^v, f_2^v, f_3^v, f_4^v\} = \mathcal{E}_{\mathrm{DINO}}\left(I_{\mathrm{din}}\right),
    \label{eq:dino_features}
\end{equation}
where $\mathcal{E}_{\mathrm{DINO}}(\cdot)$ denotes the frozen DINOv3 encoder, and $f_m^v$ represents the visual feature at the $m$-th level. DINOv3 features from different depths contain multi-granularity visual information ranging from local structures to high-level context, making them suitable for providing hierarchical visual priors to the multi-scale restoration network.

Since the channel dimensions and spatial resolutions of the original DINOv3 features are inconsistent with those of the restoration network, we design a lightweight prior adapter to perform spatial alignment and channel mapping. For the $m$-th visual feature $f_m^v$, it is first resized to the spatial size of the corresponding restoration scale via bilinear interpolation, and then reorganized by an adaptation module composed of a $1\times1$ convolution, a depthwise separable convolution, and a projection layer:
\begin{equation}
    v_m = G_m\left(\mathrm{Resize}\left(f_m^v\right)\right),
    \quad m=1,\ldots,4,
    \label{eq:visual_prior_adapter}
\end{equation}
where $G_m(\cdot)$ denotes the visual prior adaptation module corresponding to the $m$-th scale, and $v_m \in \mathbb{R}^{C_v \times H_m \times W_m}$ represents the visual prior feature aligned with the corresponding scale of the restoration network. Accordingly, the multi-scale visual prior set is obtained as
\begin{equation}
    \mathcal{V} = \{v_1, v_2, v_3, v_4\}.
    \label{eq:visual_prior_set}
\end{equation}

\subsubsection{Visual Prior Modulation} 
After visual prior adaptation, directly concatenating DINOv3 features with restoration features may allow the external visual priors to participate indiscriminately in the entire restoration process, thereby weakening the constraint imposed by observation consistency modeling on the input degraded image. To address this issue, we restrict these priors to the prior variable update branch and adopt a residual gated modulation mechanism, enabling them to participate in the reconstruction process in an explicit and controllable manner, as shown in Figure~\ref{fig:de-visual}(b).

Let $h_l^{(k)}$ denote the intermediate feature for prior restoration at the $l$-th scale in the $k$-th unfolding stage, and let $v_l$ denote the visual prior at the corresponding scale. First, a visual prior modulation projection module is used to generate spatially varying scaling and shifting terms from $v_l$:
\begin{equation}
    [\gamma_l^{(k)}, \beta_l^{(k)}] = M_l^{(k)}(v_l),
    \label{eq:visual_prior_modulation_params}
\end{equation}
where $M_l^{(k)}(\cdot)$ denotes the visual prior modulation projection module, and $\gamma_l^{(k)}$ and $\beta_l^{(k)}$ represent the scale and shift parameters generated from the visual prior, respectively. Then, conditional modulation is performed on the prior restoration feature:
\begin{equation}
    h_{\mathrm{sft},l}^{(k)} =
    h_l^{(k)} \odot \left(1+\gamma_l^{(k)}\right) + \beta_l^{(k)},
    \label{eq:visual_prior_sft}
\end{equation}
where $\odot$ denotes element-wise multiplication. To prevent the visual prior from introducing overly strong perturbations to the restoration features at the early training stage, we further introduce a learnable residual modulation strength $\lambda_l^{(k)}$ and obtain the modulated feature through residual gating:
\begin{equation}
    \hat{h}_l^{(k)} =
    h_l^{(k)} + \rho_l^{(k)}
    \left(h_{\mathrm{sft},l}^{(k)} - h_l^{(k)}\right),
    \label{eq:residual_gated_modulation}
\end{equation}
where $\rho_l^{(k)}$ is initialized to $0$, allowing the model to approximate a prior restoration module without DINOv3 modulation at the early training stage. As training proceeds, the network can adaptively learn the injection strength of the visual prior. This strategy gradually introduces the multi-level visual constraints provided by DINOv3 while maintaining training stability.

In implementation, the above modulation mechanism is embedded into the Transformer blocks of the prior restoration module. Each block first updates the prior restoration feature through a self-attention module, then performs residual gated modulation using the visual prior at the corresponding scale, and finally reorganizes the feature through a feed-forward network. Let $P_l(\cdot; v_l)$ denote the restoration operator with visual prior modulation. The predicted prior variable can be written as
\begin{equation}
    \tilde{z}_l^{(k+1)} =
    P_l\left(x_l^{(k+1)}; v_l\right),
    \label{eq:visual_prior_guided_prediction}
\end{equation}
where $\tilde{z}_l^{(k+1)}$ denotes the prior state predicted by the prior restoration module according to the updated data variable $x_l^{(k+1)}$ and the visual prior $v_l$. Subsequently, a residual update strategy is adopted to progressively inject the predicted prior variable into the current prior variable:
\begin{equation}
    z_l^{(k+1)} =
    z_l^{(k)} + \alpha_l^{(k)}
    \left(\tilde{z}_l^{(k+1)} - z_l^{(k)}\right),
    \label{eq:residual_prior_update}
\end{equation}
where $\alpha_l^{(k)}$ denotes the stage-dependent learnable update weight. Compared with directly replacing $z_l^{(k)}$ with $\tilde{z}_l^{(k+1)}$, the residual update provides a smoother optimization path and reduces the risk of instability caused by excessive intervention from the visual prior.

Through the above design, the visual-prior-guided reconstruction branch restricts the hierarchical visual priors provided by DINOv3 to the prior variable update process. Without directly interfering with degradation-aware observation consistency modeling, this branch enhances the recovery of structural details in locally damaged or severely degraded regions.
%

\section{Experiments and analysis}
\subsection{Experimental Setup} 
\subsubsection{Datasets}
To comprehensively evaluate the image restoration capability of DVANet, we conduct experiments on single-degradation, complex nighttime degradation, composite degradation, and cross-domain restoration tasks. For single-degradation tasks, we evaluate image desnowing performance on Snow100K~\cite{snow100k}, SRRS~\cite{jstasr}, and CSD~\cite{csd}; in-distribution and out-of-distribution performance on the denoising task is evaluated on BSD68~\cite{bsd68}; image dehazing performance is evaluated on Dense-Haze\cite{dense-haze} and NH-Haze\cite{nh-haze}; and low-light image enhancement performance is evaluated on LOL-V2 Real and LOL-V2 Synthetic~\cite{lol-v2}. For the complex nighttime degradation task, we conduct experiments on the HQ-NightRain~\cite{cst} dataset, which contains three subsets: RS, RD, and SD. For composite degradation tasks, we evaluate DVANet on the LOLBlur~\cite{lednet} and CDD11~\cite{onerestore} datasets. In addition, to further evaluate cross-domain adaptation capability, we conduct experiments on the medical image dataset CEC~\cite{endouic} and the remote sensing image datasets Haze1K~\cite{satehaze1k} and RICE~\cite{rs-cloud}.

\subsubsection{Implementation Details} 
During training, we employ a frozen DINOv3 ViT-L/16 to extract features from the 6th, 12th, 18th, and 24th layers as visual priors. Following previous methods~\cite{focalnet,convir,bioir}, we optimize the model using a dual-domain L1 loss. For the denoising task, images are cropped into $128 \times 128$ patches with a batch size of 32, and the model is trained for 150 epochs with an initial learning rate of $2 \times 10^{-4}$. For the challenging nighttime deraining task, images are cropped into $256 \times 256$ patches with a batch size of 8, and the model is trained for 200K iterations. For other restoration tasks, the model is generally trained for 400K iterations. All experiments are conducted on four NVIDIA Tesla A100 40GB GPUs.
%

\begin{table}[tb]
\centering
\caption{Quantitative comparison on image denoising on the BSD68~\cite{bsd68} dataset. PSNR (dB, $\uparrow$) is reported. Noise levels $\sigma=15,25,50$ are treated as in-distribution (ID), while $\sigma=60,75$ are treated as out-of-distribution (OOD).}
\label{tab:denoise}
\setlength{\tabcolsep}{4.5pt}
\renewcommand\arraystretch{1.10}
\definecolor{oursbg}{HTML}{F2FDE8}
\resizebox{\linewidth}{!}{
\begin{tabular}{l|c|ccc|cc}
\hline
\multirow{2}{*}{Method} 
& \multirow{2}{*}{Venue}
& \multicolumn{3}{c|}{ID} 
& \multicolumn{2}{c}{OOD} \\
\cline{3-7}
& 
& $\sigma=15$ & $\sigma=25$ & $\sigma=50$ 
& $\sigma=60$ & $\sigma=75$ \\
\hline
SwinIR~\cite{swinir} & ICCV'21 & 33.31 & 30.59 & 27.13 & 24.39 & 20.11 \\
NAFNet~\cite{nafnet} & ECCV'22 & 33.67 & 31.02 & 27.73 & \underline{25.90} & 19.42 \\
IR-SDE~\cite{ir-sde} & ICML'23 & 33.14 & 30.46 & 26.98 & 17.55 & 13.35 \\
RAM~\cite{ram} & ECCV'24 & 33.63 & 31.06 & 27.80 & 24.95 & 20.82 \\
DA-CLIP~\cite{da-clip} & ICLR'24 & \underline{34.05} & 31.20 & 27.85 & 19.68 & 16.92 \\
DiffUIR~\cite{diffuir} & CVPR'24 & 33.86 & 30.88 & 26.63 & 22.25 & 18.89 \\
DCPT~\cite{dcpt} & ICLR'25 & 33.82 & 31.16 & 27.86 & 24.49 & 20.29 \\
DFPIR~\cite{dfpir} & CVPR'25 & 33.94 & 31.29 & 28.05 & 25.85 & \underline{21.28} \\
DA-RCOT~\cite{da-rcot} & TPAMI'25 & 33.84 & 30.91 & 25.95 & 21.62 & 16.29 \\
MoCE-IR~\cite{moceir} & CVPR'25 & 34.00 & \underline{31.34} & \underline{28.07} & 24.89 & 20.12 \\
\hline
\rowcolor{oursbg}
\textbf{DVANet} 
& \textbf{Ours}
& \textbf{34.24} 
& \textbf{31.64} 
& \textbf{28.41} 
& \textbf{26.43} 
& \textbf{21.31} \\
\hline
\end{tabular}}
\vspace{-1em}
\end{table}
%

\begin{table}[tb]
\centering
\caption{Quantitative comparison on image desnowing across CSD~\cite{csd}, SRRS~\cite{jstasr}, and Snow100K~\cite{snow100k} datasets.}
\label{tab:desnow}
\setlength{\tabcolsep}{5.0pt}
\renewcommand\arraystretch{1.10}
\definecolor{oursbg}{HTML}{F2FDE8}
\resizebox{\linewidth}{!}{\begin{tabular}{l|cc|cc|cc}
\hline
\multirow{2}{*}{Methods} & \multicolumn{2}{c|}{CSD} & \multicolumn{2}{c|}{SRRS} & \multicolumn{2}{c}{Snow100K} \\
\cline{2-7}
& PSNR$\uparrow$ & SSIM$\uparrow$ & PSNR$\uparrow$ & SSIM$\uparrow$ & PSNR$\uparrow$ & SSIM$\uparrow$ \\
\hline
DesnowNet~\cite{snow100k} & 20.13 & 0.81 & 20.38 & 0.84 & 30.50 & 0.94 \\
CycleGAN~\cite{cyclegan} & 20.98 & 0.80 & 20.21 & 0.74 & 26.81 & 0.89 \\
JSTASR~\cite{jstasr} & 27.96 & 0.88 & 25.82 & 0.89 & 23.12 & 0.86 \\
HDCW-Net~\cite{csd} & 29.06 & 0.91 & 27.78 & 0.92 & 31.54 & 0.95 \\
TransWeather~\cite{transweather} & 31.76 & 0.93 & 28.29 & 0.92 & 31.82 & 0.93 \\
NAFNet~\cite{nafnet} & 33.13 & 0.96 & 29.72 & 0.94 & 32.41 & 0.95 \\
FocalNet~\cite{focalnet} & 37.18 & 0.99 & 31.34 & 0.98 & 33.53 & 0.95 \\
MSPFormer~\cite{mspformer} & 33.75 & 0.96 & 30.76 & 0.95 & 33.43 & 0.96 \\
MBTF-V1~\cite{mbtf-v1} & - & - & \underline{32.26} & 0.98 & 33.79 & 0.96 \\
IRNeXt~\cite{irnext} & \underline{37.29} & \underline{0.99} & 31.91 & 0.98 & 33.61 & 0.95 \\
PEUNet~\cite{peunet} & 37.28 & 0.97 & 31.89 & \textbf{0.98} & \underline{34.11} & \underline{0.96} \\
\hline
\rowcolor{oursbg}
\textbf{DVANet} & \textbf{38.16} & \textbf{0.99} & \textbf{32.81} & 0.97 & \textbf{34.33} & \textbf{0.96} \\
\hline
\end{tabular}}
\end{table}
%

\begin{table}[tb]
\centering
\caption{Quantitative comparison on image dehazing across Dense-Haze~\cite{dense-haze} and NH-Haze~\cite{nh-haze} datasets.}
\label{tab:dehaze}
\setlength{\tabcolsep}{4.2pt}
\renewcommand\arraystretch{1.15}
\definecolor{oursbg}{HTML}{F2FDE8}
\resizebox{\linewidth}{!}{\begin{tabular}{l|c|cc|cc}
\hline
\multirow{2}{*}{Method} & \multirow{2}{*}{Venue} & \multicolumn{2}{c|}{Dense-Haze} & \multicolumn{2}{c}{NH-Haze} \\
\cline{3-6}
& & PSNR$\uparrow$ & SSIM$\uparrow$ & PSNR$\uparrow$ & SSIM$\uparrow$ \\
\hline
TransWeather~\cite{transweather} & CVPR'22 & 10.51 & 0.452 & 11.58 & 0.411 \\
Uformer~\cite{uformer} & CVPR'22 & 15.22 & 0.430 & - & - \\
PromptIR~\cite{promptir} & NeurIPS'23 & 9.57 & 0.433 & 11.38 & 0.434 \\
DA-CLIP~\cite{da-clip} & ICLR'24 & 10.94 & 0.459 & 12.35 & 0.466 \\
AutoDIR~\cite{autodir} & ECCV'24 & 12.33 & \underline{0.486} & 12.71 & 0.477 \\
X-Restormer~\cite{x-restormer} & ECCV'24 & 9.57 & 0.430 & 11.36 & 0.413 \\
InstructIR~\cite{instructir} & ECCV'24 & 11.03 & 0.465 & 12.24 & 0.498 \\
DiffUIR~\cite{diffuir} & CVPR'24 & 9.59 & 0.433 & 11.39 & 0.422 \\
FoundIR~\cite{foundir} & ICCV'25 & 9.29 & 0.431 & 11.43 & 0.449 \\
AgenticIR~\cite{agenticir} & ICLR'25 & 10.11 & 0.388 & 12.20 & 0.450 \\
FoundIR-v2~\cite{foundir-v2} & CVPR'26 & \underline{15.29} & \textbf{0.506} & \underline{17.00} & \underline{0.462} \\
\hline
\rowcolor{oursbg}
\textbf{DVANet} & Ours & \textbf{16.15} & 0.470 & \textbf{17.99} & \textbf{0.677} \\
\hline
\end{tabular}}
\end{table}
%

\begin{table}[tb]
\centering
\caption{Quantitative comparison on low-light image enhancement on the LOL-v2~\cite{lol-v2} dataset.}
\label{tab:lowlight}
\setlength{\tabcolsep}{4.2pt}
\renewcommand\arraystretch{1.15}
\definecolor{oursbg}{HTML}{F2FDE8}
\resizebox{\linewidth}{!}{\begin{tabular}{l|c|cc|cc}
\hline
\multirow{2}{*}{Method} & \multirow{2}{*}{Venue} & \multicolumn{2}{c|}{LOL-v2-Real} & \multicolumn{2}{c}{LOL-v2-Syn}\\
\cline{3-6}
& & PSNR$\uparrow$ & SSIM$\uparrow$ & PSNR$\uparrow$ & SSIM$\uparrow$ \\
\hline
DRBN~\cite{drbn} & CVPR'20 & 20.29 & 0.831 & 23.22 & 0.927 \\
MIRNet~\cite{mirnet} & ECCV'20 & 20.02 & 0.820 & 21.94 & 0.876 \\
URetinex~\cite{uretinex} & CVPR'22 & 20.44 & 0.806 & 24.73 & 0.897 \\
Uformer~\cite{uformer} & CVPR'22 & 18.82 & 0.771 & 19.66 & 0.871 \\
Restormer~\cite{restormer} & CVPR'22 & 19.94 & 0.827 & 21.41 & 0.830 \\
Retinexformer~\cite{retinexformer} & ICCV'23 & \underline{22.80} & 0.840 & 25.67 & 0.930 \\
DiffIR~\cite{diffir} & ICCV'23 & 21.15 & 0.816 & 24.76 & 0.921 \\
FourLLIE~\cite{fourllie} & MM'24 & 21.60 & 0.847 & 24.17 & 0.917 \\
FourierDiff~\cite{fourierdiff} & CVPR'24 & 18.67 & 0.602 & 13.70 & 0.631 \\
AST~\cite{ast} & CVPR'24 & 21.68 & \underline{0.856} & 22.25 & 0.927 \\
URetinexNet++~\cite{uretinexnet++} & TPAMI'25 & 21.97 & 0.836 & 24.60 & 0.927 \\
VIVNet~\cite{vivnet} & TPAMI'26 & - & - & \underline{26.01} & \underline{0.948} \\
\hline
\rowcolor{oursbg}
\textbf{DVANet} & Ours & \textbf{23.14} & \textbf{0.861} & \textbf{26.51} & \textbf{0.953} \\
\hline
\end{tabular}}
\end{table}

\begin{figure*}[!tp] 
	\centerline{\includegraphics[page=1,trim = 0mm 0mm 0mm 0mm, clip, width=1\linewidth]{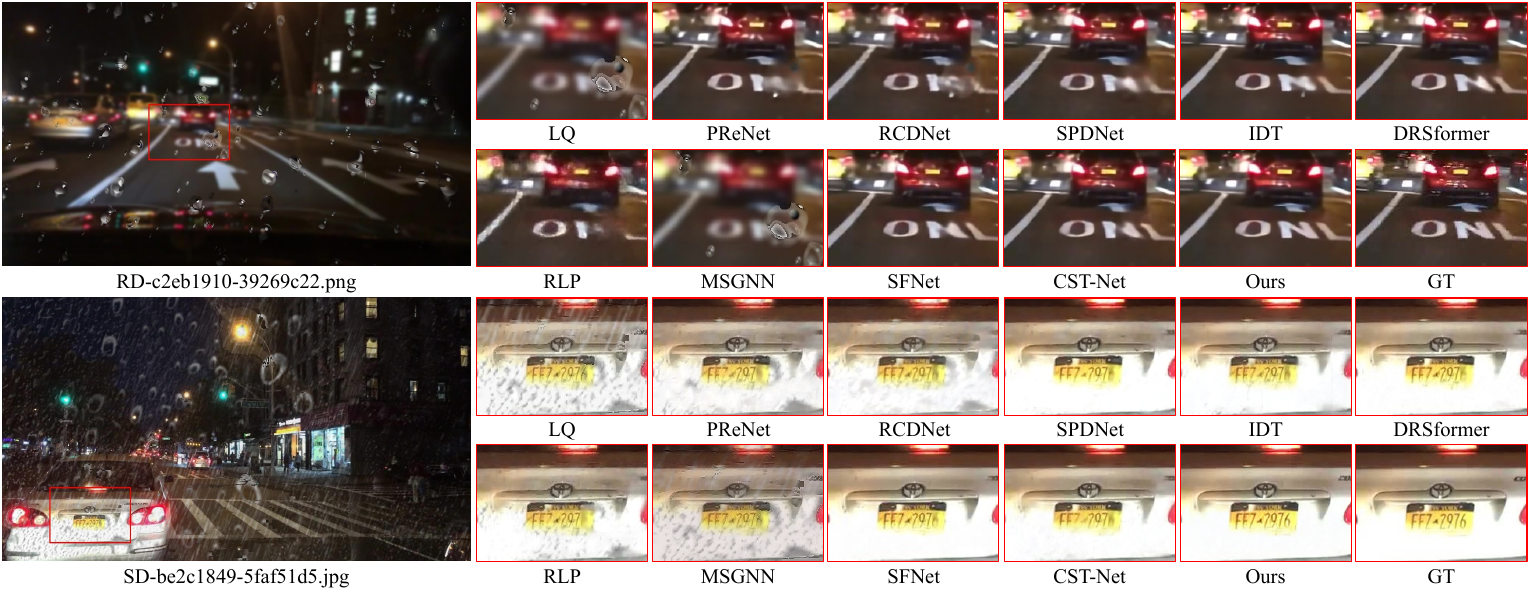}}
	\caption{
    Visual comparison of DVANet with state-of-the-art methods on the HQ-NightRain dataset, including the raindrop (RD) subset and the rain-streak-and-raindrop (SD) subset. Zoom in for the best view.}
	\label{fig:hq_nightrain}
	\vspace{-1em}
\end{figure*}
%

\begin{table}[t]
\centering
\caption{Quantitative comparison on complex nighttime image restoration on the HQ-NightRain~\cite{cst} dataset.}
\label{tab:hq_nightrain}
\setlength{\tabcolsep}{3.2pt}
\renewcommand\arraystretch{1.10}
\definecolor{oursbg}{HTML}{F2FDE8}
\resizebox{\linewidth}{!}{
\begin{tabular}{l|cc|cc|cc}
\hline
\multirow{2}{*}{Method} 
& \multicolumn{2}{c|}{RS} 
& \multicolumn{2}{c|}{RD} 
& \multicolumn{2}{c}{SD} \\
\cline{2-7}
& PSNR$\uparrow$ & SSIM$\uparrow$ 
& PSNR$\uparrow$ & SSIM$\uparrow$ 
& PSNR$\uparrow$ & SSIM$\uparrow$ \\
\hline
PReNet~\cite{prenet} & 38.585 & 0.9842 & 32.003 & 0.9432 & 34.837 & 0.9698 \\
RCDNet~\cite{rcdnet} & 37.754 & 0.9788 & 31.892 & 0.9357 & 32.780 & 0.9512 \\
SPDNet~\cite{spdnet} & 40.049 & 0.9865 & 31.477 & 0.9361 & 39.037 & 0.9825 \\
IDT~\cite{idt} & 42.420 & 0.9918 & 33.618 & 0.9522 & 38.487 & 0.9843 \\
Restormer~\cite{restormer} & 41.884 & 0.9907 & 33.796 & 0.9503 & 40.179 & 0.9884 \\
SFNet~\cite{sfnet} & 41.481 & 0.9920 & 33.606 & 0.9465 & 40.301 & 0.9875 \\
DRSformer~\cite{drsformer} & 42.811 & 0.9922 & 33.845 & 0.9491 & 40.432 & \underline{0.9886} \\
RLP~\cite{rlp} & 40.409 & 0.9885 & 29.973 & 0.9204 & 31.130 & 0.9709 \\
MSGNN~\cite{msgnn} & 27.718 & 0.8846 & 24.515 & 0.8244 & 27.634 & 0.9078 \\
NeRD-Rain~\cite{nerd-rain} & 42.714 & 0.9923 & 33.831 & 0.9500 & 39.683 & 0.9855 \\
CST-Net~\cite{cst} & \underline{42.885} & \underline{0.9924} & \underline{33.940} & \underline{0.9523} & \underline{40.498} & 0.9881 \\
\hline
\rowcolor{oursbg}
\textbf{DVANet} & \textbf{46.143} & \textbf{0.9968} & \textbf{34.108} & \textbf{0.9702} & \textbf{43.823} & \textbf{0.9955} \\
\hline
\end{tabular}}
\end{table}
%

\begin{table*}[t]
\centering
\caption{Quantitative comparison on composite degradation restoration on the LoLBlur~\cite{lednet} dataset.}
\label{tab:lolblur}
\setlength{\tabcolsep}{3.0pt}
\renewcommand{\arraystretch}{1.12}
\definecolor{oursbg}{HTML}{F2FDE8}
\resizebox{\textwidth}{!}{\begin{tabular}{c|ccccccccccccc}
\hline
Method & DeblurGAN-v2~\cite{deblurgan-v2} & DRBN~\cite{drbn} & MIMO~\cite{mimo} & NAFNet~\cite{nafnet} & LEDNet~\cite{lednet} & Restormer~\cite{restormer} & Retinexformer~\cite{retinexformer} & DarkIR~\cite{darkir} & DSwinIR~\cite{dswinir} & VIVNet~\cite{vivnet} & \cellcolor{oursbg}\textbf{DVANet} \\
Venue & ICCV'19 & CVPR'20 & ICCV'21 & ECCV'22 & ECCV'22 & CVPR'22 & ICCV'23 & CVPR'25 & TPAMI'25 & TPAMI'26 & \cellcolor{oursbg}Ours \\
\hline
PSNR$\uparrow$ & 22.30 & 21.78 & 22.41 & 25.36 & 25.74 & 26.72 & 26.02 & 27.30 & \underline{27.33} & 27.20 & \cellcolor{oursbg}\textbf{27.52} \\
SSIM$\uparrow$ & 0.745 & 0.768 & 0.745 & 0.882 & 0.850 & 0.902 & 0.887 & 0.898 & \underline{0.909} & 0.903 & \cellcolor{oursbg}\textbf{0.923} \\
\hline
\end{tabular}}
\end{table*}
%

\begin{table*}[t]
\centering
\caption{Quantitative comparison on the CDD11~\cite{onerestore} dataset under single, double, and triple degradation settings. PSNR (dB, $\uparrow$) is reported.}
\label{tab:cdd11}
\setlength{\tabcolsep}{3.2pt}
\renewcommand\arraystretch{1.12}
\definecolor{oursbg}{HTML}{F2FDE8}
\resizebox{\textwidth}{!}{
\begin{tabular}{l|c|cccc|ccccc|cc|c}
\hline
\multirow{2}{*}{Method} & \multirow{2}{*}{Venue} & \multicolumn{4}{c|}{\textit{CDD11-Single}} & \multicolumn{5}{c|}{\textit{CDD11-Double}} & \multicolumn{2}{c|}{\textit{CDD11-Triple}} & \multirow{2}{*}{Average} \\
\cline{3-13}
& & Low (L) & Haze (H) & Rain (R) & Snow (S) & L+H & L+R & L+S & H+R & H+S & L+H+R & L+H+S & \\
\hline
NAFNet~\cite{nafnet} & ECCV'22 & 24.50 & 25.34 & 33.10 & 33.43 & 24.80 & 25.28 & 24.99 & 26.80 & 26.15 & 23.90 & 23.82 & 26.99 \\
AirNet~\cite{airnet} & CVPR'22 & 24.83 & 24.21 & 26.55 & 26.79 & 23.23 & 22.82 & 23.29 & 22.21 & 23.29 & 21.80 & 22.24 & 23.75 \\
TransWeather~\cite{transweather} & CVPR'22 & 23.39 & 23.95 & 26.69 & 25.74 & 22.24 & 22.62 & 21.80 & 23.10 & 22.34 & 21.55 & 21.01 & 23.13 \\
PromptIR~\cite{promptir} & NeurIPS'23 & 26.32 & 26.10 & 31.56 & 31.53 & 24.49 & 25.05 & 24.51 & 24.54 & 23.70 & 23.74 & 23.33 & 25.90 \\
WGWSNet~\cite{wgwsnet} & CVPR'23 & 24.39 & 27.90 & 33.15 & 34.43 & 24.27 & 25.06 & 24.60 & 27.23 & 27.65 & 23.90 & 23.97 & 26.96 \\
WeatherDiff~\cite{weatherdiff} & TPAMI'23 & 23.58 & 21.99 & 24.85 & 24.80 & 21.83 & 22.69 & 22.12 & 21.25 & 21.99 & 21.23 & 21.04 & 22.49 \\
OneRestore~\cite{onerestore} & ECCV'24 & 26.48 & 32.52 & 33.40 & 34.31 & 25.79 & 25.58 & 25.19 & \underline{29.99} & \underline{30.21} & 24.78 & 24.90 & 28.47 \\
MoCE-IR~\cite{moceir} & CVPR'25 & \textbf{27.26} & \underline{32.66} & \underline{34.31} & \underline{35.91} & \textbf{26.24} & \textbf{26.25} & \underline{26.04} & 29.93 & 30.19 & \textbf{25.41} & \underline{25.39} & 29.05 \\
MIRAGE~\cite{mirage} & ICLR'26 & \underline{27.13} & 32.39 & 34.23 & 35.57 & \underline{26.04} & 26.21 & \textbf{26.07} & 29.49 & 29.72 & \underline{25.17} & \textbf{25.41} & 28.86 \\
\hline
\rowcolor{oursbg}
\textbf{DVANet} & \textbf{Ours} & 26.07 & \textbf{36.39} & \textbf{35.12} & \textbf{37.17} & 25.31 & \underline{26.23} & 24.44 & \textbf{32.08} & \textbf{32.80} & 24.97 & 22.80 & \textbf{29.40} \\
\hline
\end{tabular}}
\end{table*}
%

\subsection{One-by-One Image Restoration} 
To verify the modeling capability of DVANet for different single degradation types, we conduct independent training and testing on image denoising, desnowing, dehazing, and low-light enhancement tasks, respectively. For each task, the model is trained separately on the corresponding dataset to evaluate the restoration capability of DVANet under specific degradation conditions.

\subsubsection{Image Denoising} 
Table~\ref{tab:denoise} presents the quantitative results on the image denoising task. DVANet achieves superior denoising performance under different noise levels. In particular, under the out-of-distribution noise setting, DVANet improves upon MoCE-IR~\cite{moceir} by 1.54 dB and 1.19 dB at $\sigma=60$ and $\sigma=75$, respectively, indicating its stronger generalization ability to unseen severe noise.

\subsubsection{Image Desnowing} 
Table~\ref{tab:desnow} reports the image desnowing results. DVANet achieves the highest PSNR on all three datasets. Compared with the best competing methods, DVANet improves the PSNR by 0.87 dB, 0.55 dB, and 0.22 dB on the CSD, SRRS, and Snow100K datasets, respectively. In terms of SSIM, DVANet also achieves the best or near-best performance. These results demonstrate that DVANet can well adapt to different snow degradation scenarios and maintain stable restoration performance under various degradation conditions.

\subsubsection{Image Dehazing} 
Table~\ref{tab:dehaze} shows the comparison results. Different from regular homogeneous haze, these two datasets contain more complex haze distributions and local degradation variations. DVANet enhances the adaptive modeling capability for haze degradation through global-local degradation representation and degradation-conditioned data mapping and further improves structural detail recovery via visual-prior-guided reconstruction. As a result, DVANet achieves better quantitative performance in complex dehazing scenarios.

\subsubsection{Low-Light Image Enhancement} 
Table~\ref{tab:lowlight} presents the quantitative comparison results on the low-light image enhancement task. DVANet achieves PSNR values of 23.14 dB and 26.51 dB on the LOL-v2-Real and LOL-v2-Syn datasets, respectively, outperforming the recent method URetinexNet++\cite{uretinexnet++} by 1.17 dB and 1.91 dB. These results indicate that DVANet can effectively suppress noise in dark regions and preserve structural details while improving the visibility of low-light images, thereby achieving superior enhancement performance.
%

\begin{figure*}[!tp] 
	\centerline{\includegraphics[page=1,trim = 0mm 0mm 0mm 0mm, clip, width=1\linewidth]{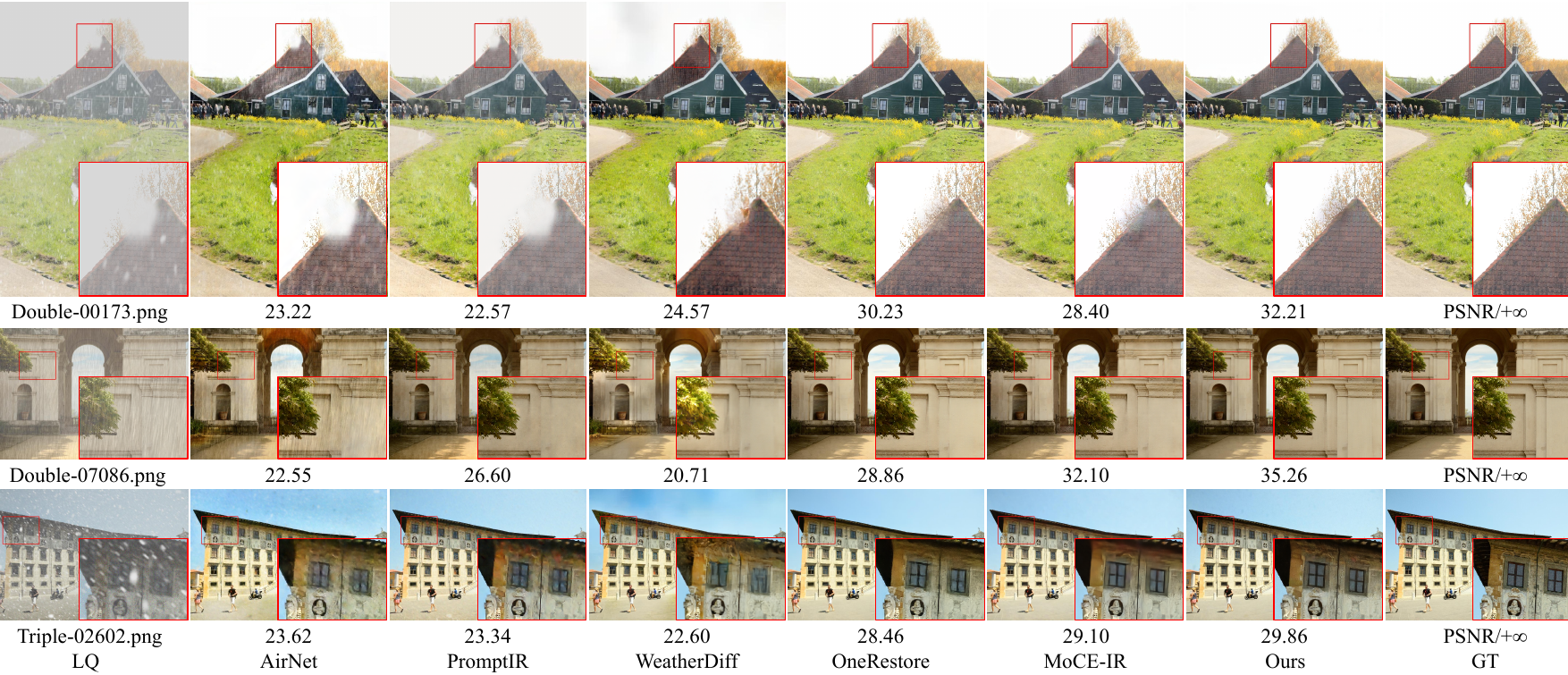}}
	\caption{Visual comparison of DVANet with state-of-the-art methods on the CDD11 dataset under composite degradation settings. The first, second, and third rows show results for haze+snow, haze+rain, and low-light+haze+snow degradation, respectively. Zoom in for the best view.}
	\label{fig:cdd-11}
	\vspace{-1em}
\end{figure*}
%

\begin{table*}[t]
\centering
\caption{Quantitative comparison on medical image restoration on the CEC~\cite{endouic} dataset.}
\label{tab:cec}
\setlength{\tabcolsep}{3.0pt}
\renewcommand{\arraystretch}{1.12}
\definecolor{oursbg}{HTML}{F2FDE8}
\resizebox{\textwidth}{!}{\begin{tabular}{c|ccccccccccccc}
\hline
Method & PromptIR~\cite{promptir} & LLCaps~\cite{llcaps} & Diff-LOL~\cite{diff-lol} & LA-Net~\cite{la-net} & LighTDiff~\cite{lightdiff} & EndoUIC~\cite{endouic} & AMIR~\cite{amir} & MoCE-IR~\cite{moceir} & DFPIR~\cite{dfpir} & EndoIR~\cite{endoir} & \cellcolor{oursbg}\textbf{DVANet} \\
Venue & NeurIPS'23 & MICCAI'23 & TOG'23 & IJCV'23 & MICCAI'24 & MICCAI'24 & MICCAI'24 & CVPR'25 & CVPR'25 & AAAI'26 & \cellcolor{oursbg}Ours \\
\hline
PSNR$\uparrow$ & 28.27 & 27.55 & 28.07 & 15.86 & 29.23 & 26.39 & 32.27 & 29.44 & 26.34 & \underline{32.61} & \cellcolor{oursbg}\textbf{38.64} \\
SSIM$\uparrow$ & 0.831 & 0.860 & 0.961 & 0.664 & 0.961 & 0.956 & 0.964 & 0.933 & 0.842 & \underline{0.975} & \cellcolor{oursbg}\textbf{0.997} \\
\hline
\end{tabular}}
\end{table*}
%

\begin{table*}[t]
\centering
\caption{Quantitative comparison on remote sensing image declouding on the RS-Cloud~\cite{rs-cloud} dataset.}
\label{tab:rice}
\setlength{\tabcolsep}{3.0pt}
\renewcommand{\arraystretch}{1.12}
\definecolor{oursbg}{HTML}{F2FDE8}
\resizebox{\textwidth}{!}{\begin{tabular}{c|ccccccccccccc}
\hline
Method & TransWeather~\cite{transweather} & PromptIR~\cite{promptir} & DA-CLIP~\cite{da-clip} & AutoDIR~\cite{autodir} & X-Restormer~\cite{x-restormer} & InstructIR~\cite{instructir} & DiffUIR~\cite{diffuir} & AgenticIR~\cite{agenticir} & FoundIR-v2~\cite{foundir-v2} & TPGDiff~\cite{tpgdiff} & \cellcolor{oursbg}\textbf{DVANet} \\
Venue & CVPR'22 & NeurIPS'23 & ICLR'24 & ECCV'24 & ECCV'24 & ECCV'24 & CVPR'24 & ICLR'25 & CVPR'26 & ICML'26 & \cellcolor{oursbg}Ours \\
\hline
PSNR$\uparrow$ & 12.93 & 11.35 & 16.43 & 18.39 & 11.48 & 14.46 & 13.82 & 17.80 & 22.06 & \underline{38.37} & \cellcolor{oursbg}\textbf{38.54} \\
SSIM$\uparrow$ & 0.708 & 0.772 & 0.803 & 0.809 & 0.751 & 0.865 & 0.814 & 0.799 & 0.828 & \underline{0.954} & \cellcolor{oursbg}\textbf{0.990} \\
\hline
\end{tabular}}
\end{table*}
%

\begin{table}[tb]
\centering
\caption{Quantitative comparison on remote sensing image dehazing on the SateHaze1K~\cite{satehaze1k} dataset.}
\label{tab:haze1k}
\setlength{\tabcolsep}{4.5pt}
\renewcommand\arraystretch{1.10}
\definecolor{oursbg}{HTML}{F2FDE8}
\resizebox{\linewidth}{!}{\begin{tabular}{l|c|cc|cc}
\hline
\multirow{2}{*}{Method} & \multirow{2}{*}{Venue} & \multicolumn{2}{c|}{Thin} & \multicolumn{2}{c}{Thick} \\
\cline{3-6}
& & PSNR$\uparrow$ & SSIM$\uparrow$ & PSNR$\uparrow$ & SSIM$\uparrow$ \\
\hline
AOD-Net~\cite{aod-net} & ICCV'17 & 19.54 & 0.854 & 15.92 & 0.731 \\
H$^2$RL-Net~\cite{h2rl-net} & GRSL'21 & 20.91 & 0.880 & 17.41 & 0.768 \\
FCTF-Net~\cite{fctf-net} & TGRS'20 & 23.59 & 0.913 & 20.03 & 0.816 \\
Uformer~\cite{uformer} & CVPR'22 & 22.82 & 0.907 & 20.36 & 0.815 \\
C$^2$PNet~\cite{c2pnet} & CVPR'23 & 19.62 & 0.880 & 16.83 & 0.790 \\
Restormer~\cite{restormer} & CVPR'22 & 23.08 & 0.912 & 18.58 & 0.762 \\
UMWTrans~\cite{umwtran} & ECCV'22 & 24.29 & \underline{0.919} & 20.07 & 0.825 \\
FocalNet~\cite{focalnet} & ICCV'23 & 24.16 & 0.916 & 21.69 & 0.847 \\
DehazeFormer~\cite{dehazeformer} & TIP'23 & 24.26 & 0.909 & 22.26 & 0.835 \\
EMPF~\cite{empf} & TGRS'23 & 22.69 & 0.896 & 20.23 & 0.822 \\
Trinity~\cite{trinity} & TGRS'23 & 22.65 & 0.896 & 20.57 & 0.824 \\
FMambaIR~\cite{fmambair} & TGRS'25 & \underline{24.58} & 0.912 & \underline{22.65} & \underline{0.850} \\
\hline
\rowcolor{oursbg}
\textbf{DVANet} & Ours & \textbf{25.94} & \textbf{0.928} & \textbf{22.95} & \textbf{0.859} \\
\hline
\end{tabular}}
\end{table}
%

\begin{figure}[!tbp]  
	\centerline{\includegraphics[page=1,trim = 0mm 0mm 0mm 0mm, clip, width=1\linewidth]{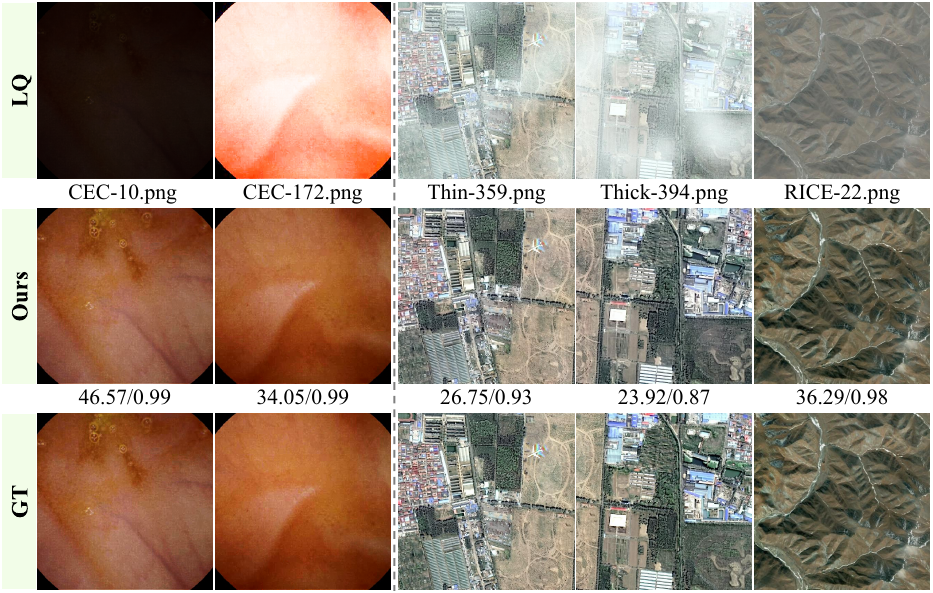}}
	\caption{Visual comparison of DVANet on cross-domain image restoration tasks, including medical and remote sensing image restoration.}
	\label{fig:cross-domain}
	\vspace{-1em}
\end{figure}
%

\begin{table}[tb]
\centering
\caption{Ablation study on the effectiveness of degradation and visual priors in DVANet.}
\label{tab:ablation_components}
\setlength{\tabcolsep}{3.8pt}
\renewcommand\arraystretch{1.10}
\definecolor{oursbg}{HTML}{F2FDE8}
\resizebox{\linewidth}{!}{
\begin{tabular}{c|cc|cc|cc|cc}
\hline
\multirow{2}{*}{Setting} & \multirow{2}{*}{Deg.} & \multirow{2}{*}{Vis.} & \multicolumn{2}{c|}{RainDrop} & \multicolumn{2}{c|}{Smoke} & \multicolumn{2}{c}{Avg.} \\
\cline{4-9}
& & 
& PSNR$\uparrow$ & SSIM$\uparrow$ 
& PSNR$\uparrow$ & SSIM$\uparrow$ 
& PSNR$\uparrow$ & SSIM$\uparrow$ \\
\hline
Base & \xmark & \xmark & 31.22 & 0.943 & 27.82 & 0.876 & 29.52 & 0.910 \\
w/o Deg. & \xmark & \cmark & 31.52 & 0.945 & 28.34 & 0.881 & 29.93 & 0.913 \\
w/o Vis. & \cmark & \xmark & 31.39 & 0.944 & 28.06 & 0.879 & 29.73 & 0.912 \\
\rowcolor{oursbg}
Ours & \cmark & \cmark & \textbf{31.66} & \textbf{0.946} & \textbf{28.51} & \textbf{0.882} & \textbf{30.09} & \textbf{0.914} \\
\hline
\end{tabular}}
\end{table}

%

\begin{table}[tb]
\centering
\caption{Ablation study on global and local degradation representations in DVANet.}
\label{tab:ablation_degradation}
\setlength{\tabcolsep}{3.8pt}
\renewcommand\arraystretch{1.10}
\definecolor{oursbg}{HTML}{F2FDE8}
\resizebox{\linewidth}{!}{
\begin{tabular}{c|cc|cc|cc|cc}
\hline
\multirow{2}{*}{Setting} 
& \multirow{2}{*}{$\mathbf{d}_{\mathrm{gl}}$} 
& \multirow{2}{*}{$\mathbf{d}_{\mathrm{lo}}$} 
& \multicolumn{2}{c|}{RainDrop} 
& \multicolumn{2}{c|}{Smoke} 
& \multicolumn{2}{c}{Avg.} \\
\cline{4-9}
& & 
& PSNR$\uparrow$ & SSIM$\uparrow$ 
& PSNR$\uparrow$ & SSIM$\uparrow$ 
& PSNR$\uparrow$ & SSIM$\uparrow$ \\
\hline
Base & \xmark & \xmark & 31.22 & 0.943 & 27.82 & 0.876 & 29.52 & 0.910 \\
w/o $\mathbf{d}_{\mathrm{gl}}$ & \xmark & \cmark & 31.28 & 0.943 & 27.91 & 0.877 & 29.60 & 0.910 \\
w/o $\mathbf{d}_{\mathrm{lo}}$ & \cmark & \xmark & 31.34 & 0.944 & 28.00 & 0.877 & 29.67 & 0.911 \\
\rowcolor{oursbg}
Ours & \cmark & \cmark & \textbf{31.39} & \textbf{0.944} & \textbf{28.06} & \textbf{0.879} & \textbf{29.73} & \textbf{0.912} \\
\hline
\end{tabular}}
\end{table}
%

\begin{figure}[!tbp]  
	\centerline{\includegraphics[page=1,trim = 0mm 0mm 0mm 0mm, clip, width=1\linewidth]{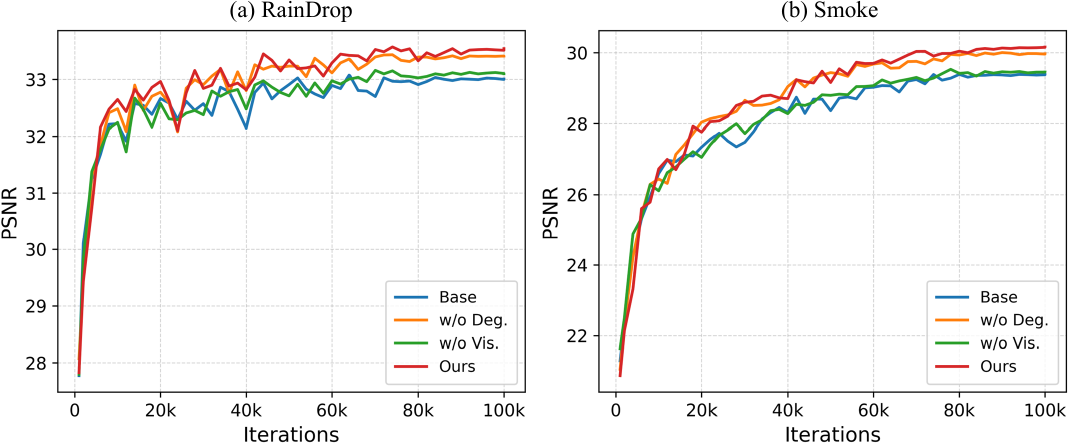}}
	\caption{Validation PSNR curves of different prior ablation variants on the RainDrop and Smoke datasets.}
	\label{fig:Abla-Com}
	\vspace{-1em}
\end{figure}
%

\subsection{Complex Nighttime Image Restoration} 
To further evaluate the restoration capability of DVANet in complex nighttime degradation scenarios, we conduct experiments on the HQ-NightRain~\cite{cst} dataset. Compared with conventional image restoration tasks, nighttime rain images are usually affected by multiple factors, including low illumination, local highlights, rain streaks, and raindrops, posing higher requirements on the model's degradation adaptability, local interference suppression capability, and structural detail recovery ability. As shown in Table~\ref{tab:hq_nightrain}, DVANet achieves the best performance on all three subsets, demonstrating its effectiveness in adapting to complex nighttime rain degradation scenarios. The visual results in Figure~\ref{fig:hq_nightrain} show that DVANet can effectively reduce nighttime rain degradation residues, suppress local highlight interference, and recover clearer structural edges and detailed textures.
%

\subsection{Composite Degradation Image Restoration} 
To validate the restoration capability of DVANet in complex composite degradation scenarios, we further evaluate it on composite degradation image restoration tasks. As shown in Table~\ref{tab:lolblur}, DVANet achieves superior results on the LOLBlur~\cite{lednet} dataset, indicating its effectiveness in handling composite degradations where low-light conditions and blur coexist. Furthermore, we conduct evaluations on the more challenging CDD11~\cite{onerestore} dataset, which contains rain, haze, snow, low-light degradation, and their various combinations, forming 11 restoration settings, with each image containing up to three degradation types simultaneously. As shown in Table~\ref{tab:cdd11}, compared with OneRestore, DVANet improves the average PSNR by 0.93 dB, demonstrating stronger robustness and generalization ability in composite degradation scenarios. As shown in Figure~\ref{fig:cdd-11}, DVANet more effectively removes multiple degradation residues and restores clearer structural edges and more natural local details.

\subsection{Cross-Domain Image Restoration} 
To further verify the cross-domain generalization capability of DVANet, we apply it to two specialized scenarios: medical image restoration and remote sensing image restoration. Compared with natural images, medical images and remote sensing images exhibit significant differences in imaging mechanisms, texture distributions, structural patterns, and degradation modes, making them suitable for evaluating the model's adaptability to data from different domains.

\subsubsection{Medical Image Restoration} 
As shown in Table~\ref{tab:cec}, DVANet achieves superior or competitive performance on medical image restoration tasks. As shown in the left two columns of Figure~\ref{fig:cross-domain}, while suppressing low-light degradation, DVANet restores tissue brightness and local texture distributions that are closer to the GT and better preserves the continuity of low-contrast structures, further verifying its cross-domain adaptability in medical image restoration tasks.

\subsubsection{Remote Sensing Image Restoration} 
As shown in Tables~\ref{tab:rice} and~\ref{tab:haze1k}, DVANet also exhibits strong competitiveness on remote sensing image restoration tasks. Remote sensing images usually contain structured regions such as roads, buildings, vegetation, and water bodies and thus have high requirements for edge continuity and regional consistency. As shown in the right three columns of Figure~\ref{fig:cross-domain}, after removing degradations, DVANet better preserves road edges, building region contours, and mountainous texture structures, indicating that its degradation-aware updating mechanism and visual-prior restoration mechanism exhibit certain cross-domain transfer ability.
%

\subsection{Ablation Study} 
To validate the effectiveness of the key designs in DVANet, we conduct systematic ablation experiments on image raindrop removal and image smoke removal tasks and train the models on the RainDrop~\cite{raindrop} and Smoke~\cite{smoke} datasets for 100K iterations, respectively.

\subsubsection{ Effects of Different Components}
As shown in Table~\ref{tab:ablation_components}, compared with the baseline model, introducing either degradation-aware information or visual priors alone improves the restoration performance. Specifically, degradation-aware information provides input-dependent degradation conditions for observation consistency modeling, while visual priors provide structural and semantic information for the prior variable update. Furthermore, the training curves in Figure~\ref{fig:Abla-Com} show that the full model achieves higher and more stable PSNR on both the RainDrop and Smoke datasets, validating the effectiveness of each component in improving restoration performance and promoting stable convergence.
%

\subsubsection{Effects of Degradation Representation} 
We further compare the effects of different degradation representation strategies. As shown in Table~\ref{tab:ablation_degradation}, using only global degradation information can enhance the model's perception of the overall degradation state, but its ability to characterize local degradation variations is limited. Using only local degradation information can describe region-level degradation cues but lacks global degradation context. Compared with using a single representation, jointly using global and local degradation information achieves the best performance, indicating that complex degradation modeling requires considering both image-level degradation trends and region-level degradation variations.
%

\subsection{Discussion and Limitation} 
Although DVANet achieves good performance on various image restoration tasks, there are still several directions worthy of further investigation. First, although the frozen DINOv3~\cite{dinov3} visual encoder provides effective structural and contextual priors and helps maintain regional consistency, it also introduces additional computational and storage costs. Therefore, how to design a more lightweight and efficient visual prior extraction and injection mechanism remains an important problem for future exploration. Second, this paper mainly focuses on static image restoration and can be further extended to video restoration, cross-modal restoration, and real-world degradation modeling in open scenarios in the future. 
%

\section{Conclusion} 
This paper investigates unified image restoration under complex degradation conditions and proposes DVANet, an optimization-inspired dual-variable unfolding framework guided by degradation awareness and visual priors. By integrating degradation-aware observation consistency with visual-prior-guided reconstruction, DVANet can compensate for corrupted structural information while maintaining consistency with degraded observations. Extensive experiments across single-degradation, complex nighttime degradation, composite degradation, and cross-domain restoration scenarios demonstrate the strong adaptability and robustness of DVANet, verifying the effectiveness of the proposed collaborative unfolding strategy.
%



%
%

\bibliographystyle{IEEEtran}
\bibliography{trans_ref}
\end{document}